\documentclass[twocolumn, switch]{article} %
\usepackage{preprint}

\usepackage[utf8]{inputenc} %
\usepackage[T1]{fontenc}    %
\usepackage{hyperref}       %
\usepackage{url}            %
\usepackage{booktabs}       %
\usepackage{amsfonts}       %
\usepackage{nicefrac}       %
\usepackage{microtype}      %
\usepackage{lipsum}		%
\usepackage{graphicx}
\usepackage{lineno}		%
\usepackage{float}			%

\usepackage{subfigure}

\usepackage{mlapa}

\usepackage{multirow}
\usepackage{threeparttable}

\usepackage{empheq}

\usepackage{color}

\usepackage{mathtools}

\usepackage{titlesec}
\titlespacing\section{0pt}{12pt plus 3pt minus 3pt}{1pt plus 1pt minus 1pt}
\titlespacing\subsection{0pt}{10pt plus 3pt minus 3pt}{1pt plus 1pt minus 1pt}
\titlespacing\subsubsection{0pt}{8pt plus 3pt minus 3pt}{1pt plus 1pt minus 1pt}

\title{Comparison of Syntactic Parsers on Biomedical Texts}
\usepackage{authblk}

\author[1,2\thanks{\tt{maria.biryukov@uni.lu}}]{Maria Biryukov}

\affil[1]{Luxembourg Center for Contemporary and Digital History, University of Luxembourg}
\affil[2]{Luxembourg Center for Systems Biomedicine, University of Luxembourg}

\begin{document}

\twocolumn[ %
  \begin{@twocolumnfalse} %
  \maketitle
\begin{abstract}

Syntactic parsing is an important step in the automated text analysis which aims at information extraction. Quality of the syntactic parsing determines to a large extent the recall and precision of the text mining results. In this paper we evaluate the performance of several popular syntactic parsers in application to the biomedical text mining.\footnotemark       

\end{abstract}
\vspace{0.35cm}

  \end{@twocolumnfalse} %
] %
\footnotetext{Most of this work was done in 2018 when the author was primarily affiliated with LCSB, Bioinformatic Core.}

\section{Introduction and related work}
Biomedical information extraction from text is an active area of research with applications to disease maps generation, construction of protein-protein interaction networks, and more. The backbone of such applications are semantic relations between relevant concepts, and this relational information is encoded in syntactic dependencies between the words representing the concepts. It means that proper understanding of the syntactic structure of a text is a precondition for correct interpretation of its meaning. 
In the last years many high quality syntactic parsers became publicly available leading to a natural question of choice of the best and most appropriate one for a specific task. The question calls for parser evaluation and comparison. 

To date, parsers have been evaluated from a variety of perspectives. McDonald and Nivre~\cite{McDonaldNivre:2011} compared two prevailing dependency parsing paradigms, transition and graph-based, with the goal to connect error types with the theoretical aspect of each model. For the accuracy evaluation they consider sentence length, dependency distance, valency (the number of node's siblings), part-of-speech and main dependency types: root, subject and object. Recently Choi~\cite{ItDepends} compared $10$ state-of-the-art parsers in terms of their accuracy and speed in application to a number of genres covering newswire, broadcast news and conversation, web text and telephone conversation. 

The most important metrics used in the parser evaluation CoNLL shared tasks are accuracies of unlabeled (UAS) and labeled (LAS) word attachments. The former check correctness of words' syntactic head assignment while the latter controls also correctness of dependency label between the words' pairs. Although these metrics give a feeling of overall parser performance, they do not shed light on how errors are propagated and affect parse components. Kummerfeld et. al.~\cite{ParserShowdown} implemented an alternative evaluation procedure aiming to estimate the downstream effect of each error. Pyysalo et. al.~\cite{PayysaloEvaluation} pointed out differences in parsing schemes which impose a challenge for reliable comparison of the results. 

Another point of interest is evaluation of parser efficiency within specific domains and tasks. In the biomedical area, Clegg and Shepherd~\cite{Clegg:Shepherd} used manually annotated abstracts from GENIA corpus~\cite{GeniaCorpus} to benchmark four parsers that have been applied in bioinformatic projects. They identified Charniak-Lease and Bikel parsers to perform the best overall and specifically on subtask of gene expression extraction. Miyao et. al.~\cite{Miyao:2008} measured performance of eight parsers used as component of an information extraction pipeline aiming at protein-protein interactions. They found that retraining with domain-specific data improves parsers accuracy although to different extent. 

Growing availability of full text articles for text mining called for comparison between article abstracts and article bodies. A few years ago Cohen et. al.~\cite{Cohen:Verspoor:2010} analyzed $97$ full text articles assembled in CRAFT corpus~\cite{craft} and evaluated performance of various text mining tools on their abstracts and bodies. They showed that article bodies have longer sentences, significantly higher usage of passives, negation and parenthesized material, compared to article bodies. Part-of-speech tagger and syntactic parser performed differently on abstracts and bodies, with the former significantly better on abstracts, while the latter insignificantly better on bodies\footnote{Syntactic parser performance was measured in terms of \textit{bracket recall} which shows how well a parser identified sentence constituent boundaries compared to gold standard.}. 

In this work we evaluate performance of seven state-of-the-art parsers on biomedical texts. We pursue the following goals: 
\begin{itemize}
\item compare parsers performance on abstracts and full texts;
\item investigate main difference between these two types of text that influence parser performance; 
\item evaluate parsers suitability for biomedical event extraction and fine-grained contextualization;
\end{itemize}

\subsection{Parsers and models for the comparison}
\label{subsec:parsers}
For this study we used seven different statistical parsers. 
Four of these are Stanford parsers\footnote{\url{https://nlp.stanford.edu/software/}} based on different linguistic paradigms: PCFG, Factored, RNN and SNN. The other three are BLLIP\footnote{\url{https://github.com/BLLIP/bllip-parser}},  McParseface\footnote{\url{https://github.com/tensorflow/models/tree/master/syntaxnet}} (referred to as Google throughout the paper), and Mate\footnote{\url{https://code.google.com/archive/p/mate-tools/wikis/ParserAndModels.wiki}}.  
The parsers are briefly introduced bellow: 
\begin{itemize}
\item  \bf{PCFG} is an accurate unlexicalized parser based on a probabilistic context-free grammar (PCFG)~\cite{KleinManning:2003}. 
\item Factored parser combines syntactic structures (PCFG) with semantic dependencies provided by using lexical information~\cite{KleinManning:2003}. Lexicalization proofed efficient when dealing with language ambiguities and helps correctly identify dependencies between sentence constituents.
\item \bf{Recursive neural networks (RNN)} parser~\cite{SocherManning:2011} works in two steps: first it uses the parses of PCFG parser to train; then recursive neural networks are trained with semantic word vectors and used to score parse trees. In this way syntactic structure and lexical information are jointly exploited.
\item \bf{Neural Network Parser (SNN)}~\cite{ChenManningSnn} is a greedy transition-based dependency parser with dense word embedding. 
\item \bf{BLLIP} is a two-stage parser composed of generative PCFG-like constituent parser and discriminative maximum entropy reranker~\cite{CharniakJohnson2005}. This parser provided the basis for various self-training experiments aiming at parser domain adaptation~\cite{McClossky},~\cite{SelfTrainPars}. 
\item \bf{Parsey McParseface} is a transition-based neural network parser with feature embeddings ~\cite{GoogleParser}. The parser performs beam search for maintaining multiple hypotheses (as opposed to the 'greedy' approach), followed by global normalization. 
\item \bf{Mate}~\cite{Mate:Bonnet2010} is a dependency parser which combines the second order maximum spanning tree and passive-aggressive perceptron algorithms to achieve high accuracy, high training and medium parsing speed.  
\end{itemize}
All the parsers are distributed with pre-trained English models, and BLLIP has in addition separate models for less formal and medical slang. The scope of genres used to train English models varies among the parsers. The most diverse is that of Stanford, and encompasses newswire, abstracts of biomedical articles found in GENIA corpus\footnote{\url{https://nlp.stanford.edu/~mcclosky/biomedical.html}}, broadcast conversations, weblogs and some technical vocabulary. BLLIP separates newswire-oriented model from that trained on reviews, question-answers, newsgroups, emails, and weblogs (SANCL2012-Uniform). Its medical model is trained on GENIA. Google is similar to Stanford but has no biomedical data, and, finally, Mate has been trained on the data provided by the CoNNL2009 shared task.\footnote{\url{http://ufal.mff.cuni.cz/conll2009-st/task-description.html}}

\section{Corpora and Data sets}
\label{sec:corpus_data}
In our experiments, we used two gold standard in-domain corpora. One was released in the framework of Genia project ~\cite{GeniaCorpus}. It consists of $2000$ Medline abstracts in the area of molecular biology.
Another corpus is known as "Colorado Richly Annotated Full Text Corpus" (CRAFT) ~\cite{craft}. It contains $67$ full text articles in a wide variety of biomedical sub-domains. Both corpora are annotated from linguistic and semantic perspectives. In this work we use a portion of linguistic annotation, which concerns part-of-speech tagging and syntactic parsing.
\subsection{Data sets}
\label{subsubsec:data_sets} 
All the studied parsers were evaluated on the abstracts from Genia, and full texts from Craft. First of all, we used the parsers as they are distributed, typically trained on generic English language (see section ~\ref{subsec:parsers} for the description). Next, we trained some of the parsers on biomedical texts, and evaluated their performance with the new models.

To obtain test and train data we used the Genia division as distributed by McClossky~\cite{McClossky}, and split Craft in a similar way. 
English models of BLLIP, Google and Mate have been evaluated on the "train and development" portions of Genia and Craft corpora. Genia model and English model of the Stanford parsers have been evaluated on "test and future use" section of Genia, and "train and development" of Craft. Similarly, Craft model was evaluated on "train and development" of Genia, and "test" set of Craft. 
The details of the corpora split for the training and evaluation tasks are shown in Table~\ref{tab:data_sets}. "Mixed" model refers to our experiment with self-training of the Google parser (see subsection~\ref{sub:training} for details).

Genia and Craft constituency trees (s-expressions) have been converted to Universal Dependencies (UD) CoNLL format using Stanford converter\footnote{\url{https://nlp.stanford.edu/software/stanford-dependencies.shtml}}. To evaluate Mate parser with its original CoNLL2009 model, we used the LTH Constituent-to-Dependency Conversion Tool\footnote{\url{http://nlp.cs.lth.se/software/treebank_converter/}} due to substantial differences between the parser output and UD format of the gold standard.

\begin{table}[t]
\caption{Training and test data sets  for parser evaluation (in thousands of sentences).}
\label{tab:data_sets}
\begin{tabular}{l|l|c|c|c}
\hline
Model & Test & $N$ of sents & Train & Dev\\
\hline\hline
\multirow{2}{*}{English\tnote{a}} &Genia-S& $12-15.5$ &- &- \\
 & Craft-B& $10-14.5$ &- &- \\\hline
\multirow{2}{*}{Genia}&Genia-S& $2.5-2.7$&$14.3$ &$1.1$\\
 & Craft-B &$10-14.5$ &- &-\\\hline
\multirow{2}{*}{Craft}&Craft-S&$0.6-0.8$ &$17$&$2.1$ \\
& Genia-B&$12-15.5$ &- &-\\\hline
\multirow{2}{*}{Mixed}&Genia-S&$2.7$ &- &-\\
& Craft-S &$0.6$ &$9.5 + 8.6$ &$1.0$\\\hline
\end{tabular}
\begin{tablenotes}
  \small 
  \item[a] Suffixes '-B' and '-S' denote big and small test sets, respectively. 
\end{tablenotes}
\end{table}

\subsection{Training}
\label{sub:training}
For re-training we selected SNN because it showed, together with BLLIP, the best performance on both corpora and was the fastest of all parsers;
Mate because it scored first of $10$ leading (as of $2015$) statistical parsers ~\cite{ItDepends}, and Google - the newest and the most accurate among the best performing parsers (as of $2016$) ~\cite{GoogleParser}. We trained the parsers following recommendations of the parsers' authors.  
The original English model of SNN was trained with word embedding suggested by ~\cite{Collobert:2011}. Given the domain orientation,
we experimented with two different word embeddings - the one pretrained on the whole PubMed (dictionary size $2.351.706$, dimensions $200$) and another one, pretrained on the whole PMC (dictionary size $2.515.686$, dimensions $200$) ~\cite{PyysaloEmbedding}, as well as without embeddings. For Google, in addition to training models based on manually annotated data, we performed a self-training experiment. We first trained the parser on Genia. Using this new model, we parsed unlabelled version of Craft, and combined half of the top-ranked produced labeled data with nearly the same amount of actually hand-labeled data from the Craft corpus. Development and test sets were selected from the manually annotated data.

\begin{table*}[t]
\caption{Overall parsers performance}
\label{tab:test-overall}
\begin{tabular}{l|l|c|c|c|c|c|c}
\hline
\multirow{3}{*}{Parser} & \multirow{3}{*}{Model} &  \multicolumn{6}{c}{Test}\\\cline{3-8}
& & \multicolumn{2}{c|}{WSJ}& \multicolumn{2}{c|}{Genia}& \multicolumn{2}{c}{Craft}\\\cline{3-8}
& & UAS& LAS& UAS& LAS& UAS& LAS\\\hline\hline
Stanford RNN& English multidomain &0.90 &-- &0.84 &0.79 &0.70 &0.66\\
Stanford PCFG & &-- & 0.86 &0.81 &0.76 &0.69 &0.64 \\
Stanford Factored &-- &-- & 0.87& 0.78& 0.73 &0.67 &0.62 \\\hline
\multirow{4}{*}{Stanford SNN}&English multidomain & 0.92 & 0.90  & 0.89& \bf{0.88} & 0.74 & 0.70\\
&Craft without embedding& & & 0.83 & 0.81 & 0.75 & 0.72\\
&Craft with PubMed embedding & & &0.83&0.81 & 0.75 & 0.72 \\
&Craft with PMC embedding & & &0.83 &0.81 &0.73 &0.70\\ \hline
\multirow{3}{*}{BLLIP}& News &-- &0.92 & 0.75&0.69 & 0.70& 0.66\\
&SANCL-2012 Uniform&0.92 &0.90 & 0.74&0.68 & 0.68& 0.62 \\
&Genia+PubMed & & & 0.90 & \bf{0.88} & 0.74 & 0.70 \\\hline
\multirow{3}{*}{Google}&English multidomain & \bf{0.95} & \bf{0.93} & 0.57 & 0.52 & 0.57 & 0.51\\
&Genia& & & \bf{0.89} & \bf{0.88} & 0.74& 0.70 \\
&Craft& & & 0.84& 0.81& 0.88& 0.86 \\
&Bio Mixed& & & 0.87 & 0.84 & 0.82 & 0.79 \\\hline
\multirow{3}{*}{Mate}&English CoNLL 2009&--&0.90& -- &-- &0.78 &0.68  \\
&Genia& & & \bf{0.89} & 0.87 &0.72 &0.68  \\
&Craft& & &0.83 &0.80 & \bf{0.87} & \bf{0.85}  \\\hline

\end{tabular}
\end{table*}

\begin{table}[t]
\caption{Accuracy of part of speech tagging}
\label{tab:pos}
\begin{tabular}{l|l|c|c}
\hline
Parser & Model &  Genia&Craft\\\hline
\hline
RNN& English multidomain &0.83 &0.71 \\
PCFG & &0.82& 0.71  \\
Factored& &0.82 & 0.71 \\\hline
\multirow{2}{*}{SNN} & English multidomain & \bf{0.98} &0.89  \\
&Craft+PubMed emb. &0.98 & 0.90 \\\hline
\multirow{3}{*}{BLLIP}& News &0.82 &0.79 \\
&SANCL-2012 Uniform&0.82 &0.77  \\
&Genia+PubMed & \bf{0.98} & 0.88  \\\hline
\multirow{3}{*}{Google}&English multidomain &0.85 &0.82 \\
&Genia& \bf{0.98}& 0.88 \\
&Craft&0.94 & \bf{0.97} \\
&Bio Mixed & 0.94&0.96  \\\hline
\multirow{3}{*}{Mate}&English CoNLL 2009&-&0.87  \\
&Genia&\bf{0.98} &0.88   \\
&Craft&0.94 &\bf{0.97}   \\\hline
\end{tabular}
\end{table}

\section{Evaluation and Discussion}
\label{sec:evaluation_discussion}
 
\subsection{Overall performance}
\label{sub:overall}
For the performance assessment, we 
adopt criteria applied by the CoNLL parser evaluation shared tasks. For each parser we report the unlabeled and labeled accuracy scores $-$ UAS and LAS, respectively. LAS
determines to what extent semantic relations between the concepts represented by the nodes could be correctly interpreted, and therefore bring us closer to the ultimate goal of the information extraction pipeline. This is the reason why we apply head-dependency criterion when evaluating parsers' performance on specific syntactic patterns (Tables ~\ref{tab:test-syn_patterns}, \ref{tab:test_preps}). 

Table~\ref{tab:test-overall} shows unlabeled and labeled scores for each parser on three different corpora: Wall Street Journal stories from section $23$ of the English Penn Treebank\footnote{Results on WSJ are quoted as reported by the parsers' authors.}, Genia\footnote{In the case of Bllip, we used pre-trained Genia+PubMed model, developed by McClossky and available with the parser.}, and Craft. 
Looking at the results we immediately notice that: a) all parsers, when run with the English model, perform better on the WSJ than on either Genia or Craft; b) parser performance on full texts (Craft) is from $10\%$ to $19\%$ lower than on abstracts when run with English or Genia models\footnote{We do not include results of Mate with CoNNL2009 model on Genia. Although all gold standard trees 
were successfully converted by the LTH converter, their representation was very much different from the parser output, making the comparison unreliable.}.  

The first observation confirms a widespread belief that performance of statistical parsers depends on how close are the training and test domains. One of the reasons for that are out-of-domain words~\cite{Lease:Charniak,AnyDomainAdap,Clegg:Shepherd} which have a higher chance to be assigned wrong part-of-speech tag and, as a consequence, make the parser generate erroneous dependency. 
CHECK WORDING: To estimate the influence of part-of-speech tagging on parser accuracy, we calculated, for all non-overlapping edges between the gold and test parses, the number of times when one or two words (end nodes) were wrongly tagged in the test set. Correlation between the part of speech error and edge mismatch varied from $10\%$ to $36\%$ depending on model. For curiosity, we calculated vocabulary overlap between GENIA and CRAFT corpora with $333,000$ most frequent English words\footnote{\url{http://norvig.com/ngrams/count_1w.txt}}. The idea was to estimate the new vocabulary parsers have learned during the training of biomedically oriented models. It turned out that $68\%$ lemmas from Craft and $47\%$ lemmas from GENIA are among these frequent English words (Table~\ref{tab:corpus_comparison}). As can be seen in Table~\ref{tab:pos}, domain acquaintance leads to up to $15\%$ improvement of the part-of-speech tagging accuracy. An exception is SNN which achieves $98\%$ already with its English multidomain model. It should not be unexpected as biomedical data makes part of the training corpus. 
In addition to the vocabulary, genre of the training corpora matters. Note that the News model of BLLIP performs slightly better than SANCL-2012 Uniform which is trained on less standard English from both, lexical and structural perspectives.

We compare parser results on entire full texts from CRAFT, and abstracts from GENIA and note much lower scores for CRAFT. On the one hand, it should not be surprising given substantial linguistic and structural differences between article abstracts and bodies demonstrated by Cohen et. al.~\cite{Cohen:Verspoor:2010} On the other hand, both belong to biomedical domain and share genre of scientific writing. McClossky et. al.~\cite{McClossky} noticed that reranking parser trained on LA Times achieved high accuracy in parsing WSJ despite, although small, differences in vocabulary and style. Unfortunately neither reranking parser, BLLIP, nor any other parser among the ones we tested demonstrate similar behavior. We try to investigate the reasons of such results in the following section.

With respect to the quantitative results, Stanford SNN trained on English model, BLLIP, trained on biomedical abstracts with GENIA reranker, Google and Mate trained by us on PubMed abstracts from Genia corpus, reach almost the same scores on GENIA ($\pm 1\%$), and  
CRAFT ($\pm 3\%$). 
Note, that while being trained on the same English corpus,   
SNN outperforms by far all other Stanford's parsers. Particularly low result of the Factored parser, which is designed to take advantage of lexical information, runs against our expectations. 

Training Craft models lets parsers reach higher scores on CRAFT, although impact of the training varies considerably between the parsers.  
Google and Mate turned to be responsive to the training, resulting in $16\%$ gain in LAS for Google and $17\%$ for Mate, compared to the results obtained with the models trained on Genia. Training SNN on Craft brought only $2\%$ improvement, which might suggest that presence of common English from a large variety of genres in the training data is indispensable for that parser even though it is intended for biomedical domain. Here our observations on variable portability of training data and methods are similar to those made by Miyao et. al.~\cite{Miyao:2008} with respect to re-training of Stanford unlexicalized parser on GENIA data. The role of embeddings in the parser training had very little but negative influence: $\approx0.5\%$ loss in performance was observed with the PubMed and $2\%$ with the PMC embeddings. It contradicts the parser's results on WSJ test corpus, where the embedding brought $\approx 0.7\%$ improvement~\cite{ChenManningSnn}. However the original word embedding was created from a vast English corpora which might have positive effect. The effect of the PMC-based embedding could probably be attributed to the quality of the word vector itself. Pyysalo et. al. suspect that it might be affected by either high proportion of noisy data in the training corpus, or aggressive trimming of the rare words performed by the \textit{word2vec} implementation used to create the vector~\cite{PyysaloEmbedding}.

\begin{table}[t]
\caption{Overview of the GENIA and CRAFT corpora}
\label{tab:corpus_comparison}
\begin{tabular}{l|l|c|c|c}
\hline
\multirow{2}{*}{Corpus} & Unique &Common  & Corpora &Sent.\\ 
& Tokens& with English & overlap & length\\\hline
\hline
GENIA & 15324 & 7257 &--&26.3  \\\hline
CRAFT & 12702 & 8733 &--&25.7  \\\hline
BOTH & -- & 4656 & 4924 &-- \\\hline
\end{tabular}
\end{table}

\begin{figure}[ht]
\vskip 0.2in
\begin{center}
\centerline{\includegraphics[width=\columnwidth]{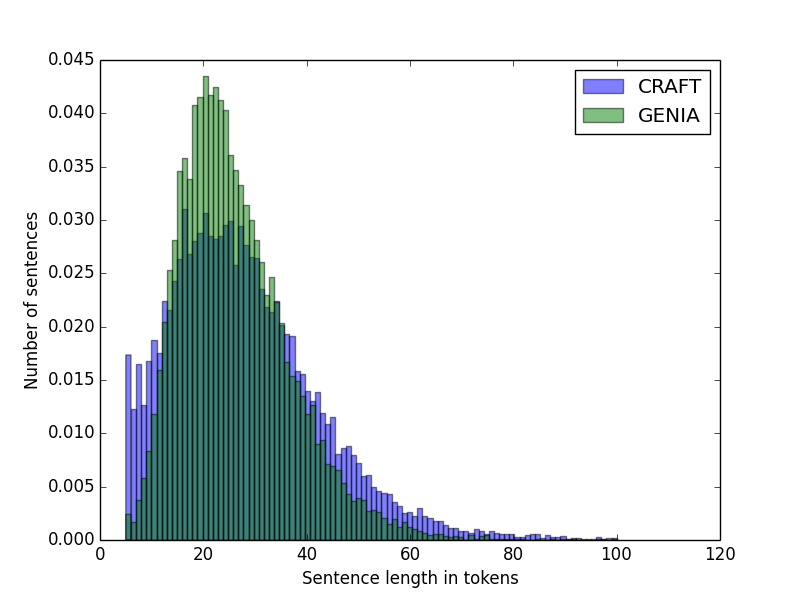}}
\caption{Sentence length distribution in GENIA and CRAFT corpora.}
\label{fig:sent_length_bins}
\end{center}
\vskip -0.2in
\end{figure}

\subsection{Corpora comparison}
\label{sub:corpus_comparison}
As already mentioned, neither common English nor English from the biomedical abstracts provide adequate training sets for parsing full texts. 

To understand what is so special about the full texts, we compare GENIA and CRAFT corpora from lexical and structural points of view.    
General corpora overview is given in Table~\ref{tab:corpus_comparison}. To calculate unique tokens we lemmatized the words, and excluded numbers and punctuation marks from computation. It turned out that the corpora share less than half of the vocabulary, and $94\%$ of that common part are from the frequent English words list. Corpus-specific items are mostly names of proteins, chemicals, mutations, coding sequences, organs and organisms. To get an idea of how challenging corpus-specific words are for overall parsing we applied Google parser to CRAFT and GENIA test sets, using Genia model for CRAFT and Craft model for GENIA\footnote{Among two parsers trained on CRAFT set, we choose Google because it showed slightly better performance than Mate.}. We then calculated impact of mistagged words on edge mismatch. It turned out that CRAFT-specific words accounted for only $8\%$ of all wrongly tagged words implicated in edge mismatch. Surprisingly, there was higher percentage of such words in the case of GENIA test set ($16\%$). However the most problematic were not the names of in-domain concepts (e.g., proteins, chemicals etc.), but words with dashed prefixes and suffixes (e.g., "pre-activation",  "gelatinase-associated" ) or words written with slash separator (e.g., "and/or" or "promoter/enhancer"). These orthographic phenomena are more diverse in GENIA than CRAFT. With the exception of molecule names and mathematical notation, dash occurrences in CRAFT are mostly limited to prefixes (e.g., "non-redundant", "re-introduce"), while slashes are found exclusively in URLs. 
Another frequent source of part-of-speech errors were 'ing'-ended words (e.g., "binding", activating"), with the confusion between gerund, adjective and noun. This is an example of syntactic ambiguity rather than corpus-specific feature. 

\begin{figure*}[t!]
    \centering
        \begin{subfigure}%
            \centering
            \includegraphics[width=\columnwidth]{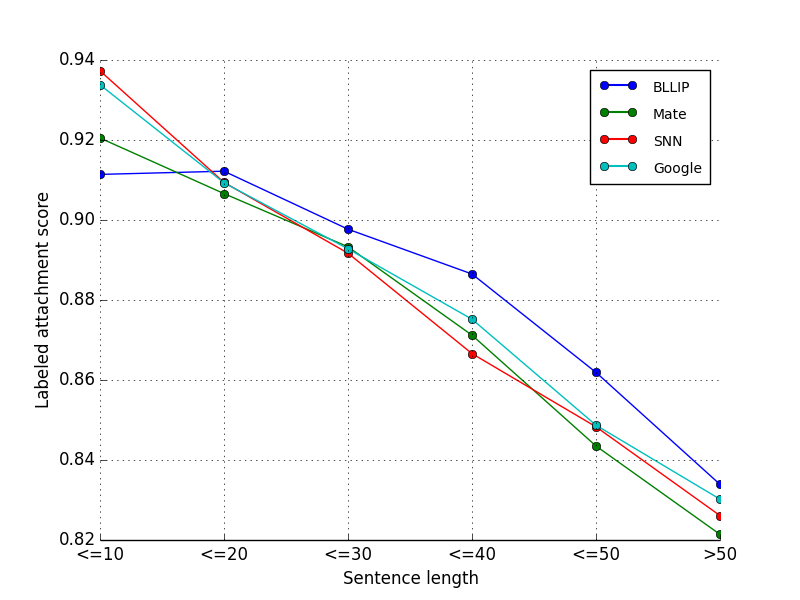}{a}
            \caption{Genia, small test set.}
            \label{subfig:genia}
        \end{subfigure}
        \begin{subfigure}%
            \centering
            \includegraphics[width=\columnwidth]{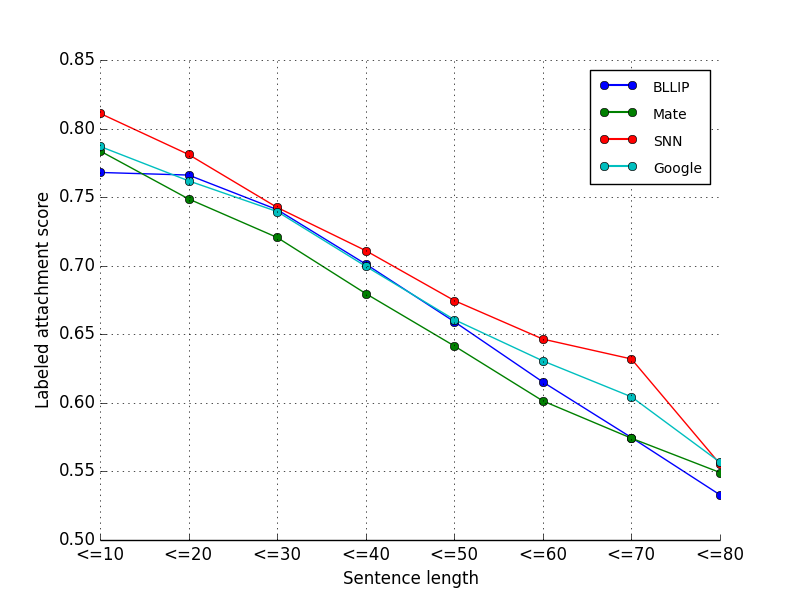}{b}
            \caption{Craft, big test set}
            \label{subfig:bigCraft}
        \end{subfigure}
        \begin{subfigure}%
            \centering
            \includegraphics[width=\columnwidth]{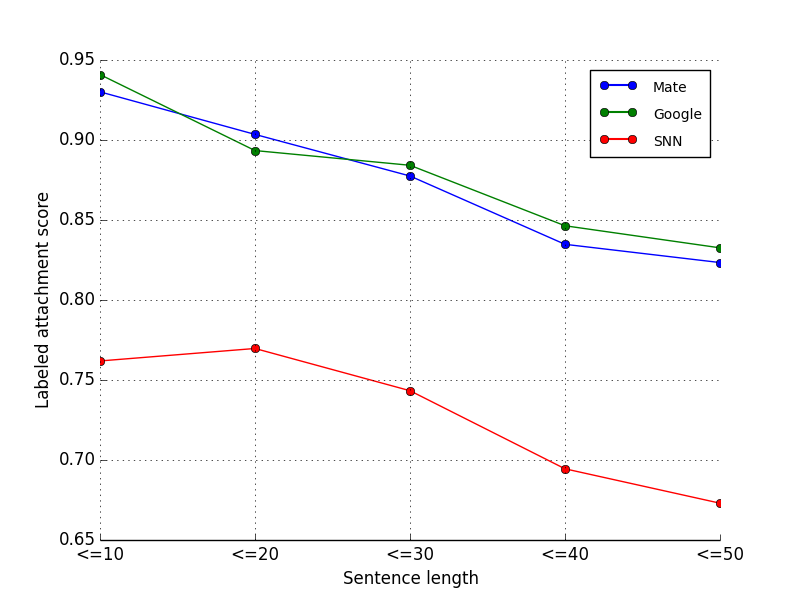}{c}
            \caption{Craft, small test set}
            \label{subfig:smallCraft}
        \end{subfigure}
    \caption{LAS versus sentence length in Genia and Craft corpora.}  
    \label{fig:las:length}
\end{figure*}
Next we looked at the structural characteristics of the corpora. Figure~\ref{fig:sent_length_bins} shows normalized sentence length distribution in GENIA and CRAFT. As one can see, CRAFT has much bigger range of sentence lengths while the average is almost the same: $26.17$ in GENIA vs $25.73$ in CRAFT (Table~\ref{tab:corpus_comparison}). Sentence length range in CRAFT is explained by the diversity of content types: captions, "Abstracts" and "Methods" sections are the shortest (mean $24.8$ and $26.8$), "Results" and "Discussion" - the longest (mean $31.17$ and $30.35$), and "Introduction" and "Conclusions" are in between (mean $26.80$ and $26.49$). 
Longer sentences have been suspected to decrease parsers performance on article bodies~\cite{Cohen:Verspoor:2010}. We measured parsers accuracies with respect to sentence lengths in bins of 10. We run the parsers with various models on GENIA and CRAFT test sets (Figure~\ref{fig:las:length},~\ref{subfig:genia} -~\ref{subfig:smallCraft}). In all combinations, scores go down as sentence length increases. On GENIA, BLLIP is the most accurate, reaching about $91\% -86\%$ LAS on the sentences up to 50 tokens. Among the remaining 3 parsers, SNN is the earliest to decline loosing $\approx3\%$ between 30 and 40 tokens length sentences, while Google and Mate do it in the next bin. On the contrary, SNN with the English multidomain model shows the best performance on CRAFT. Google and BLLIP, always with the Genia models, stick together on the sentence length range between 20 and 50 tokens, then accuracies of both continue to drop, more severely for BLLIP. Google and Mate, trained on Craft, show similar performance at every check point, reaching $\approx90-82\%$ on the length range of 20 - 50 tokens. These are nearly the same scores as they achieve on GENIA test when run with Genia models. SNN, trained on Craft, shows the same behavior but the accuracy scores remain very close to the ones obtained by the parser trained on multidomain English corpus. It confirms negligible effect of training SNN on Craft data that we have discussed in Section~\ref{sub:overall}. Importantly, all the parsers trained on Genia or multidomain English (SNN), show $15\%-20\%$ higher accuracies on GENIA than CRAFT at every bin up to 60 tokens, suggesting that sentence length is not responsible for low scores obtained by the parsers on CRAFT. Ever longer sentences are parsed poorly, but these are much less frequent even in the full texts.  %

We therefore turn to the sentence content. Manual inspection of $100$ worst scored parses produced by each tool run with Genia (or English, for SNN) models on Craft test set, showed that the most problematic texts come mainly from figure and table legends; material and procedure description from methodology sections; abbreviation lists, followed by titles and headings. Common for all these text chunks is that they often lack verb or are incomplete, i.e., they are not sentences from grammatical point of view. These are problematic for parsers irrespective of genre. For example, structurally similar pieces of text from broadcast and telephone conversations constituted major source of errors for 10 parsers evaluated by Choi et. al.~\cite{ItDepends}. Another typical feature of the low-scored sentences is abundance of mathematical notation with the whole spectrum of graphical signs, dashes and slashes. While not being parser problem per se, it affects tokenization and results in discrepancies between the test and gold standard lowing down the scores. Absence or at least under-representation of these patterns in the abstracts or general English texts may better explain decrease in parsers performance on full texts. It also suggests that successful parsing of full texts may require careful choice of tokenizing software~\cite{Diaz:Lopez:Tokenization:2015} in combination with specific training sets or methods tailored to deal with this kind of "noisy" data. It is especially important for downstream tasks aiming at extraction of technical details provided in supplementary materials, methods sections, figure and table legends.

From the practical point of view, it would be interesting to create a model which allowed a parser to perform equally well on either abstracts or full texts. Our experiment with self-training Google parser shows a promising trend. The created parser is more flexible than the ones trained on Genia or Craft individually: it performs better on Genia test set than its sibling, trained on Craft; it performs better on Craft than its sibling trained on Genia, and the difference in parsing accuracy between the Genia and Craft is also reduced. However, the absolute values leave room for training improvement.      
\begin{table*}[t]
\caption{Parsers' performance on main event-related dependencies}
\label{tab:main_deps}
\begin{tabular}{l|l|c|c|c|c|c}
\hline
\multirow{2}{*}{Parser} & \multirow{3}{*}{Corpus} & \multicolumn{5}{c}{Main Event-related Dependencies}\\\cline{3-7}
& & nsubj& nsubjpass & dobj & root & compound\\\hline
SNN + English & Genia& 0.92 & 0.93 & 0.93 & 0.94 & 0.87 \\
 & Craft & 0.78 &0.83 & 0.87 &0.86 & 0.64 \\
 SNN + Craft & Craft & 0.81 & 0.86 & 0.83 & 0.83 & 0.67\\\hline
\multirow{2}{*}{BLLIP} & Genia & \bf{0.95} & \bf{0.96} & \bf{0.95} & \bf{0.95} & \bf{0.88} \\
&Craft &0.81 & 0.87 & 0.86 & 0.89 & 0.56 \\ \hline
Google & Genia& 0.92& 0.95& 0.92 & \bf{0.95}& 0.87\\
 & Craft & 0.90 & \bf{0.96} & 0.89 & 0.92 & 0.87 \\\hline
Mate & Genia& 0.93& 0.94& 0.94& 0.95& 0.87 \\
& Craft & 0.88 & \bf{0.96} & 0.87 & 0.93 & 0.87 \\ \hline
\end{tabular}
\end{table*}

\begin{table*}[t]
\caption{Examples of syntactic patterns}
\label{tab:pattern_examples}
\begin{threeparttable}
\begin{tabular}{l|l|l}
\hline
Patterns& Examples & Explanation\\\hline\hline
PP modifying & Relief \underline{of} cyclin A gene  & Attachment of a preposition to the\\ 
& transcriptional \underline {inhibition}&  head of a prepositional phrase\\\hline
PP modified & \underline{Relief of} cyclin A gene  & Attachment of the head of a\\ &transcriptional \underline {inhibition} & prepositional phrase to the head \\ &&of the phrase it modifies\\\hline
Nominal coordination& \underline{IL1$\beta$}, \underline{IL-8} , and \underline{monocyte}  & Combining on an equal ground of\\ & \underline{chemoattractant protein}&  multiple nominal or adjectival phrases\\\hline
acl & \underline{activation triggered} by either &Clause that modifies a nominal\\ &H2O2 or TNF-$\alpha$& which precedes it\\\hline
acl:relcl&\underline{bacterium} that \underline{infects} monocytes & Clause introduced by a relative pronoun;\\& &   modifies noun which precedes it.\\\hline
advcl &treatment of K562 cells \underline{increases}  & Clause which modifies a predicate\\ &  $\gamma$-globin mRNA without \underline{affecting}  &  (verb, adjective, etc.). Cares temporal, \\ & the expression of genes...& conditional, purpose information, etc.   \\\hline
xcomp& Fos B expression is \underline{required} to &Compliment of a verb or adjective\\ & \underline{transactivate} IL-2 promoter. & without its own subject.\\\hline
\end{tabular}
\end{threeparttable}
\end{table*}

\begin{table*}[t]
\caption{Parsers' performance on specific patterns}
\label{tab:test-syn_patterns}
\begin{tabular}{l|l|c|c|c|c|c|c|c}
\hline
\multirow{2}{*}{Parser} & \multirow{2}{*}{Corpus} & \multicolumn{7}{c}{Syntactic Patterns}\\\cline{3-9}
& & PP& PP& Coordination& acl& acl:relcl&advcl& xcomp\\
& & modifying ('case')& modified ('nmod')& & & & & \\\hline\hline
SNN + & Genia& 0.94& 0.89& 0.68 &\bf{0.82}& \bf{0.81}& 0.69& 0.86\\
English&\bf{Craft}&0.85&0.82&0.54&0.71&\bf{0.73}&0.59&0.76\\\hline
\multirow{2}{*}{BLLIP} &Genia& \bf{0.95}& \bf{0.90}& \bf{0.71}& \bf{0.82}& 0.78& \bf{0.80}& \bf{0.89} \\
&Craft & 0.87 & 0.82& 0.54&0.68 &0.68 &0.66 &0.73 \\ \hline
Google & Genia& \bf{0.95}& 0.88& 0.70& 0.81& 0.76& 0.75& 0.85 \\
& \bf{Craft} & %
0.94& \bf{0.86}&\bf{0.70} &0.70 &0.68 &\bf{0.81} &\bf{0.86} \\ \hline
Mate  & Genia& 0.94& 0.87& 0.69& 0.79& 0.75& 0.70& 0.86 \\
& %
\bf{Craft} &\bf{0.95} &\bf{0.83} &0.69 &\bf{0.76} &0.64 &0.72 &0.79 \\ \hline
\end{tabular}
\end{table*}

\subsection{Parser performance on syntactic patterns}
\label{sub:patterns}
In our applications we aim at finding so-called 'events': functional connections between various biomedical concepts such as proteins, chemicals, diseases, processes. At the very basic level, event is represented by the subject-predicate-object triplet which caries information about cause, action or change of state, and theme, respectively. The most typical syntactic dependencies which encode these relations are nominal and passive subjects ('nsubj' and 'nsubjpass'), and direct object ('dobj'). These in turn are syntactic arguments of a verb\footnote{The governor of the nsubj relation is not always a verb. When the verb is a copular verb (e.g., 'be' and its inflections), semantic predicate would be expressed by an adjective or noun. E.g., in "Reduction of IL-2 production is secondary to...", the nominal subject is "Reduction", while the predicate is an adjective "secondary".} whose role is to expresses event predicate. For example, in the sentence \textit{TGF-beta reduces Ig expression.}, 'reduces' is the verb which is also event predicate. Its syntactic arguments are subject 'TGF-beta', and direct object 'Ig expression'. Their semantic roles are cause and theme respectively. Inverting this sentence into passive voice would show how 'nsubjpass' comes into play. In \textit{Ig expression is reduced by TGF-beta.}, 'Ig expression' is the passive subject of the verb 'reduced'; its second argument, 'TGF-beta', is introduced by the preposition 'by'; arguments' semantic roles do not change. 

Along with these three dependencies, we evaluate also 'root' and 'compound' relations. The former is selected because it is very often the way of introducing the main event predicate. Moreover, an error in tree root detection may come at expense of getting the whole parse wrong. With respect to 'compound' dependency, it is indispensable in multi-word entity names (e.g. in \textit{T cells receptor beta genes}, 'genes' is the head of the nominal phrase, while all the remaining words are connected to the head by 'compound' relation) which are so abundant in biomedical slang. As any nominal event argument may be expressed by such multi-word construct, accurate identification of compound relations are important for finding complete event arguments.  

The overview of the parser general performance indicates that for more accurate results it is better to use models trained on abstracts for processing the abstracts, and models trained on full texts for processing the full texts. Therefore, from now on, we evaluate the parsers with the corpus compatible models. However, since off-the-shelf state of the art parsers are trained on the abstracts, we include also results obtained by the SNN with English and BLLIP with GENIA+PubMed models on the CRAFT test set.  

Table~\ref{tab:main_deps} helps us to get a feeling of how good our parsers are for extraction of major event components. On GENIA test set, scores of all the parsers for all but 'compound' dependency are between $93\%$ and $96\%$, which is higher than their average performance. Accuracy of the 'nsubjpass' attachment is systematically higher than that of 'nsubj'. It can be explained by the presence of passive verb lexical cues (auxiliary and inflected verb, e.g., \textit{was induced}'). Compound nouns are known to be syntactically and semantically difficult to process~\cite{Cohen:nominalization}. Although this relation is labeled with lower accuracy it is about the average of parsers performance. With respect to the individual results, BLLIP is the most accurate, reaching $1\%-3\%$ more on all the dependencies except 'root', where all the parsers except SNN obtain the same score. 

On CRAFT test set, Google and Mate with Craft models, outperform BLLIP and SNN with either models. It is in line with the discussion about the overall parsers performance in Subsection~\ref{sub:overall}. Google reaches $2\%$ more on 'nsubj' and 'dobj', while Mate has $1\%$ more on 'root' dependency. Like in the case of GENIA test set, the scores here are above average for all the parsers for all the dependencies. It is also so for 'compound' in the case of Google and Mate, while SNN and especially BLLIP score poorly. We have already mentioned that training of SNN on CRAFT was much less successful compared to Google and Mate. In the context of main event-related dependencies, results reached by SNN with English and Craft models are not conclusive: Craft-trained model is better in 'nsubj', 'nsubjpass' and 'compound', while 'dobj' and 'root' are more accurate with English model. With respect to 'nsubj', 'nsubjpass' and 'dobj', BLLIP achieves (almost) the same scores as the best SNN; it outperforms either SNN model on 'root'  and is worst in 'compound' relation. With respect to 'root' dependency, SNN scores last on both test sets and with either model. It does not appear to be related to the domain orientation of our data. Choi et.a.~\cite{ItDepends} noticed the same behaviour of the parser on the OntoNotes, and attributed it to the earliest reasonable choice of root node made by the parser.     

In addition to detecting main event players, we are particularly interested in determining the event context: conditions and temporal order, coordination, mutual dependency and/or exclusion, location. Such relations are expressed via abundant use of coordination, prepositions, adverbial and relative clauses. This is the reason why, in addition to the main event-related dependencies, we look closely at parsers' performance on the corresponding syntactic patterns.

Table~\ref{tab:test-syn_patterns} shows LAS scores obtained by the selected parsers on prepositional attachment, coordination, nominal and relative clause modifier (\textit{acl} and \textit{acl:relcl}, respectively), predicate modifying clauses (\textit{advcl}), and open clausal complements (\textit{xcomp})\footnote{We do not include results of SNN trained on Craft because they were lower than those obtained by the parser run with English model for all the dependencies.}. Examples for each relation can be found in Table~\ref{tab:pattern_examples}. 

On Genia, BLLIP wins in all the dependencies except relative clause ("acl:relcl"), where SNN takes the lead. Google performs slightly better than Mate, but overall SNN, Google and Mate obtain similar results. On Craft test set, Google with Craft model shows the most stable performance, only being outperformed by Mate and SNN in nominal ("acl") and relative clause modifier assignments ("acl:relcl"). 
\begin{itemize}
\item Prepositional attachment. 
Dependencies related to prepositional attachment reach the highest scores. 'case' scores better than 'nmod' (columns 1 and 2, respectively), which could be explained by shorter dependency distance between dependent and its parent: it is twice as much for 'nmod' compared to 'case'. Overall, accuracy of the prepositional attachment is equal or above average scores for all the parsers and test sets (except Mate, where it is $2\%$ less in Craft). It is good news because prepositions are crucial in event argument detection and event contextualization. 
Table~\ref{tab:test_preps} shows parser performance on a selection of most relevant prepositions. The scores are listed as ranges, quoting results for Genia and Craft test sets. For Google and Craft, corpus-compatible models are used; for SNN and BLLIP we apply English and GENIA+PubMed models, respectively. The most accurate is the attachment of preposition 'of'. So, as far as the prepositions are concerned, unary events like \textit{inhibition of c-jun} are nicely secured. Attachment of 'in' is the worst scored. It might have negative influence on finding arguments of regulatory events (e.g., \textit{PP1 may be involved in T cell activation}) or location specification. One of the major challenges for automatic preposition processing is contextual ambiguity. Consider the phrase \textit{induction of cytokine expression in leukocytes}: what would be the correct parse: \textit{induction of [cytokine expression] in leukocytes}; \textit{[cytokine expression] in leukocytes}; both or none of the two? Syntactically speaking, the first answer is the right one.

Importantly, experiments with Google and Mate parsers show that applying corpus-compatible model considerably improves prepositional attachment accuracies. When English or Genia models are used to parse Craft test set, SNN and BLLIP obtain fairly similar results.

\begin{table}[t]
\caption{Parser performance on prepositions in \%}
\label{tab:test_preps}
\begin{tabular}{l|l|l|l|l}
\hline
Parsers & SNN& BLLIP&Google&Mate\\ %
Preps & & &\\\hline %
\bf{of} & $91-84$ & $91-82$ & $92-92$ & $91-89$\\
by & $88-80$ & $90-81$ & $87-86$ & $86-87$\\
with & $86-68$  &  $88-68$  & $87-87$  &  $84-81$ \\
to & $84-75$ & $85-72$ & $81-78$ & $81-83$ \\
between& $81-64$  &  $86-65$  &  $83-81$  &  $84-81$ \\
\bf{in} & $78-64$ & $79-67$ & $76-78$ & $76-78$\\\hline
\end{tabular}
\end{table}

\begin{figure}[ht]
\vskip 0.2in
\begin{center}
\centerline{\includegraphics[width=\columnwidth]{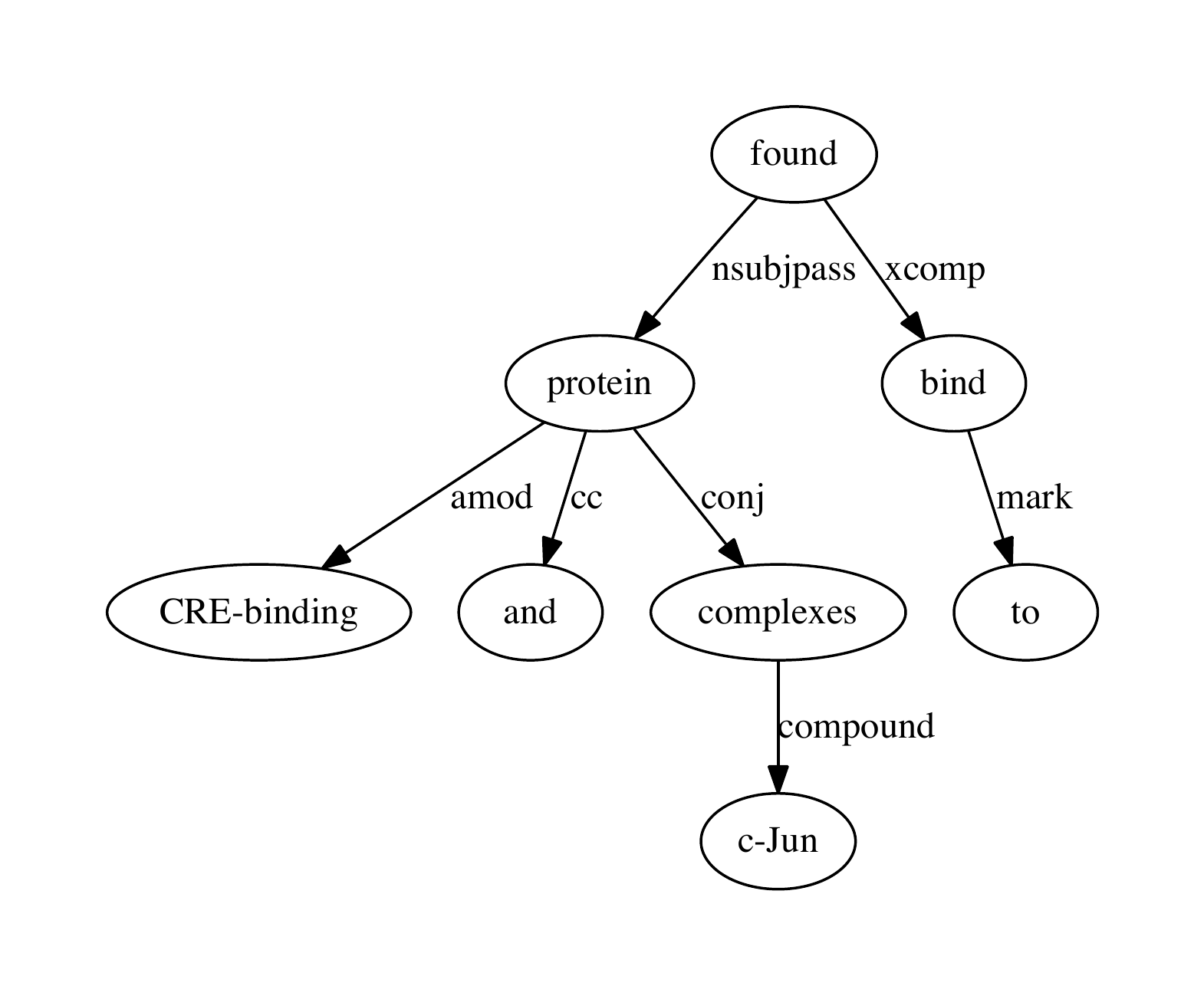}}
\caption{Coordination example. Model parse.}
\label{fig:coordination_correct}
\end{center}
\vskip -0.2in
\end{figure}

\begin{figure}[ht]
\vskip 0.2in
\begin{center}
\centerline{\includegraphics[width=\columnwidth]{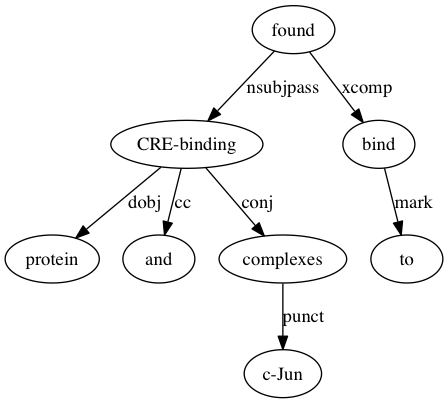}}
\caption{Coordination example (a). (Google + Craft model.)}
\label{fig:coordination_google_craft}
\end{center}
\vskip -0.2in
\end{figure}

\begin{figure}[ht]
\vskip 0.2in
\begin{center}
\centerline{\includegraphics[width=\columnwidth]{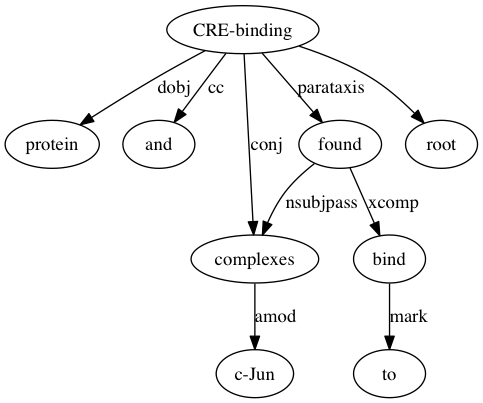}}
\caption{Coordination example (b). (Google + Mixed model)}
\label{fig:coordination_google_mixed}
\end{center}
\vskip -0.2in
\end{figure}

\begin{figure}[ht]
\vskip 0.2in
\begin{center}
\centerline{\includegraphics[width=\columnwidth]{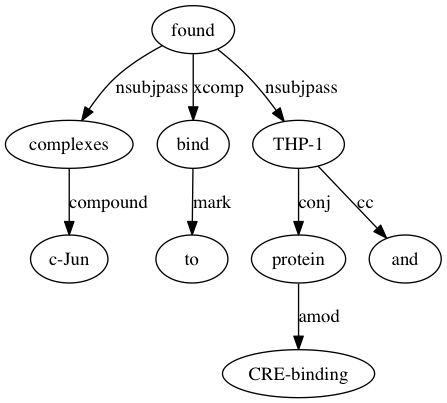}}
\caption{Coordination example (c). (Google + Genia model.)}
\label{fig:coordination_google_genia}
\end{center}
\vskip -0.2in
\end{figure}

\begin{figure}[ht]
\vskip 0.2in
\begin{center}
\centerline{\includegraphics[width=\columnwidth]{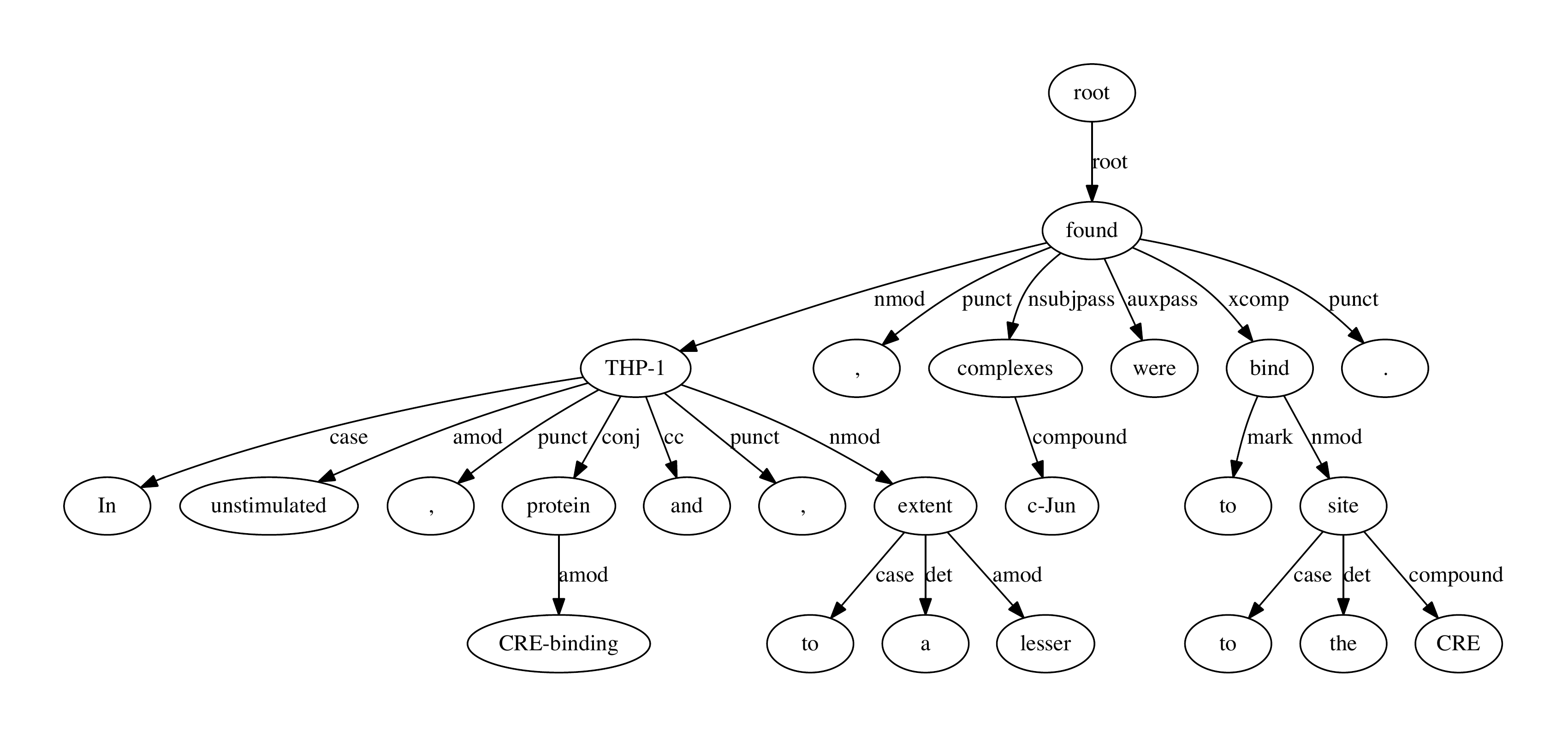}}
\caption{Coordination example (d). (Bllip + Genia and PubMed model.)}
\label{fig:coordination_Bllip}
\end{center}
\vskip -0.2in
\end{figure}

\item Coordination. Identification of the coordination scope is not easy either, often for the same reason of semantic ambiguity. In the following fragment: \textit{No differences between the normal and KA mutant mice can be detected}, are the coordinated components (1) [normal and KA mutant] mice, or (2) [normal and KA] mutant mice? The correct answer is (1) although all parsers except BLLIP yielded tree corresponding to (2). Ambiguity is not the only challenge; there are also complex constructions involving prepositions, complementary clauses and punctuation marks. In the sentence: \textit{In unstimulated THP-1, CRE-binding protein and, to a lesser extent, c-Jun complexes were found to bind to the CRE site.}, the coordinated elements are CRE-binding protein and c-Jun complexes; they participate in binding relation (Figure~\ref{fig:coordination_correct}). None of the parser-produced trees corresponds to the model; yet, from the interpretation point of view, the errors are not equally harmful. Tree in Figure~\ref{fig:coordination_google_craft} captures the coordination and scope of binding. On the contrary, the one in Figure~\ref{fig:coordination_google_genia} consider THP-1 among the coordinated elements which participate in binding. It is wrong because THP-1 is a cell line while CRE and s-Jun are proteins. The outcome based on the tree in  Figure~\ref{fig:coordination_Bllip} suggests that binding of c-Jun complexes happens in THP-1 and CRE-binding protein, which is also wrong. This example is different from the previous ones in that the sentence does not offer multiple syntactic solutions. Yet, in-domain knowledge is nearly indispensable for finding the correct parse. Note the direct object ("dobj") dependency on Figure~\ref{fig:coordination_google_mixed}, instead of the expected adjectival modifier "amod". "dobj" results from the miss-attribution of Verb to "CRE-binding". Although morphologically wrong, such reading pertains to the functional meaning captured in the protein name.

\begin{table}[t]
\caption{Most frequent confusions between modifying dependency types.}
\label{tab:dep_confusions}
\begin{tabular}{l|c}
\hline
Gold dependency & Test alternatives\\\hline\hline
acl & advcl, root, xcomp\\\hline
acl:relcl & admod, acl, ccomp \tnote{a}\\\hline
advcl & xcomp, nmod, \tnote{b} acl\\\hline
xcomp & acl, advcl\\\hline
\end{tabular}
\begin{tablenotes}
  \small 
  \item{a} ccomp: clausal complement of a verb or adjective 
  \item{b} nmod: nominal modifiers of nouns or clausal predicates 
\end{tablenotes}
\end{table}

\item Typical confusions. Remaining dependencies in Table~\ref{tab:test-syn_patterns}, 'acl', 'acl:relcl', 'advcl', 'xcomp', behave as modifiers. Table~\ref{tab:dep_confusions} shows typical edge label confusions between the gold and test parses. Naturally, the most "dangerous" swaps are the ones which wrongly identify the modified element.
Take for example, adjectival (acl) $\leftrightarrow$ adverbial (advcl) clause modifiers. The former tells us about some properties expressed by a noun, while the latter conveys details about an action expressed by a verb. Exchange of these dependencies would most probably lead to an interpretation error. Consider the following examples:
\begin{itemize}

\item \textit{acl} $\rightarrow$ \textit{advcl}. \textit{The capacity of curcumin to inhibit both cell growth and death strongly implies that ... curcumin affects a common \underline{step}, presumably \underline{involving} a modulation of the AP-1 transcription factor.} According to gold tree, what involves modulation of AP-1 TF is the \underline{step} (step$\xrightarrow{\text{acl}}$ involving). 
On the contrary, all the parsers identified \textit{affects} as dependency target of involving (affects$\xrightarrow{\text{advcl}}$ involving)
As a result, we should understand that curcumin affects a common step by means of modulation of AP-1 TF. Syntactically speaking, both parses are legal. However they lead to two different interpretations and choosing the correct one may require domain expertise.
\begin{figure}[ht]
\vskip 0.2in
\begin{center}
\centerline{\includegraphics[width=\columnwidth]{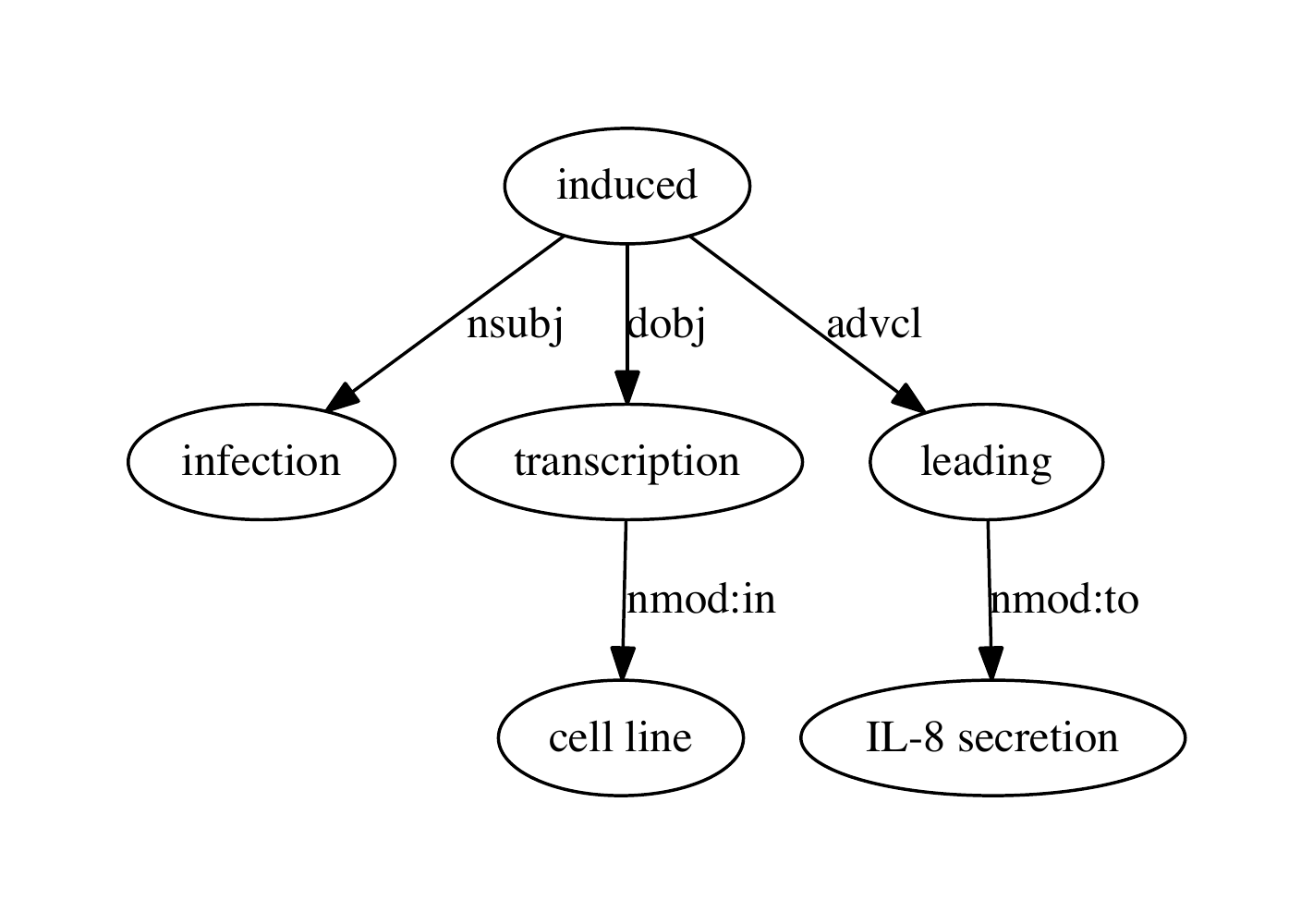}}
\caption{Adverbial clause modifier: correct (Google and Mate with Genia model.)}
\label{fig:advcl_correct}
\end{center}
\vskip -0.2in
\end{figure}
\begin{figure}[ht]
\vskip 0.2in
\begin{center}
\centerline{\includegraphics[width=\columnwidth]{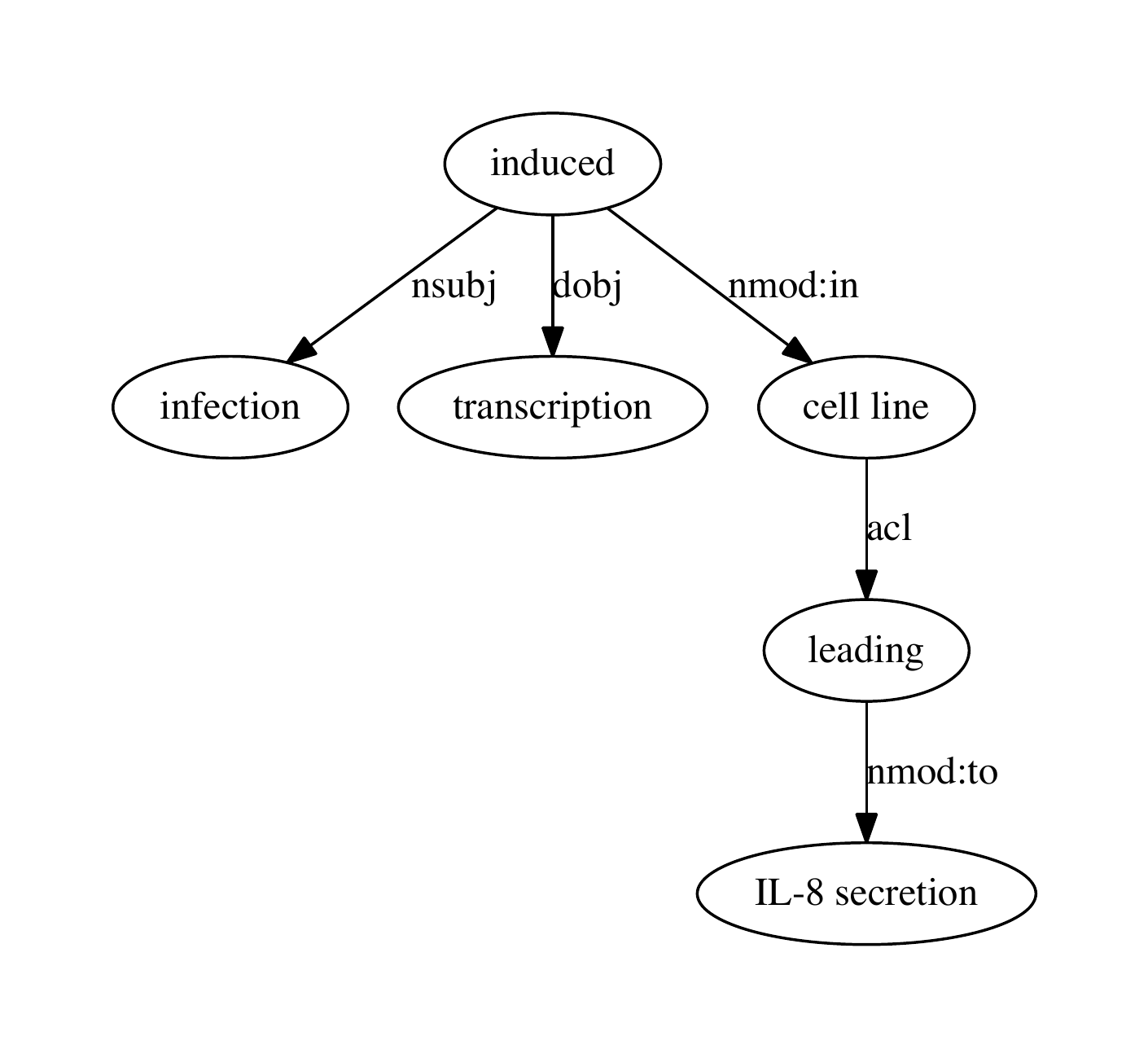}}
\caption{Adjectival clause modifier: wrong (Google with Mixed model and BLLIP)}
\label{fig:acl_wrong}
\end{center}
\vskip -0.2in
\end{figure}
\item \textit{advcl} $\rightarrow$ \textit{acl}. Adverbial and adjectival clause modifiers express often so-called "long-distance" dependencies, when the modified and modifying words are separated by many other words in the sentence. Consider an example: \textit{Cytomegalovirus infection induced interleukin-8 gene transcription in a human monocytic cell line, leading to IL-8 secretion.} When parsed correctly (Figure~\ref{fig:advcl_correct}), 'advcl' dependency should be established between induced $\xrightarrow{\text{advcl}}$ leading, thus showing that cytomegalovirus infection is the cause of IL-8 secretion. Nevertheless, some parsers suggested line $\xrightarrow{\text{acl}}$ leading edge instead~\ref{fig:acl_wrong}, making 'line' to be the cause of IL-8 secretion, which is wrong.
\item 'root'. Erroneous attribution of 'root' dependency is often an artifact resulting from parsing titles and headings, which are in many cases not real sentences. For example, in \textit{Rescue by cytokines of apoptotic cell \underline{death} \underline{induced }by IL-2 deprivation of human antigen-specific T cell clones.}, the 'acl' dependency: death $\xrightarrow{\text{acl}}$ induced, is substituted by root $\xrightarrow{\text{root}}$ induced dependency, which is wrong. It is not surprising that such substitutions occur much more often in Craft, which is composed of full texts and, in addition to article titles contains (sub)section headings and figure captions. 
\end{itemize}
\begin{figure}[ht]
\vskip 0.2in
\begin{center}
\centerline{\includegraphics[width=\columnwidth]{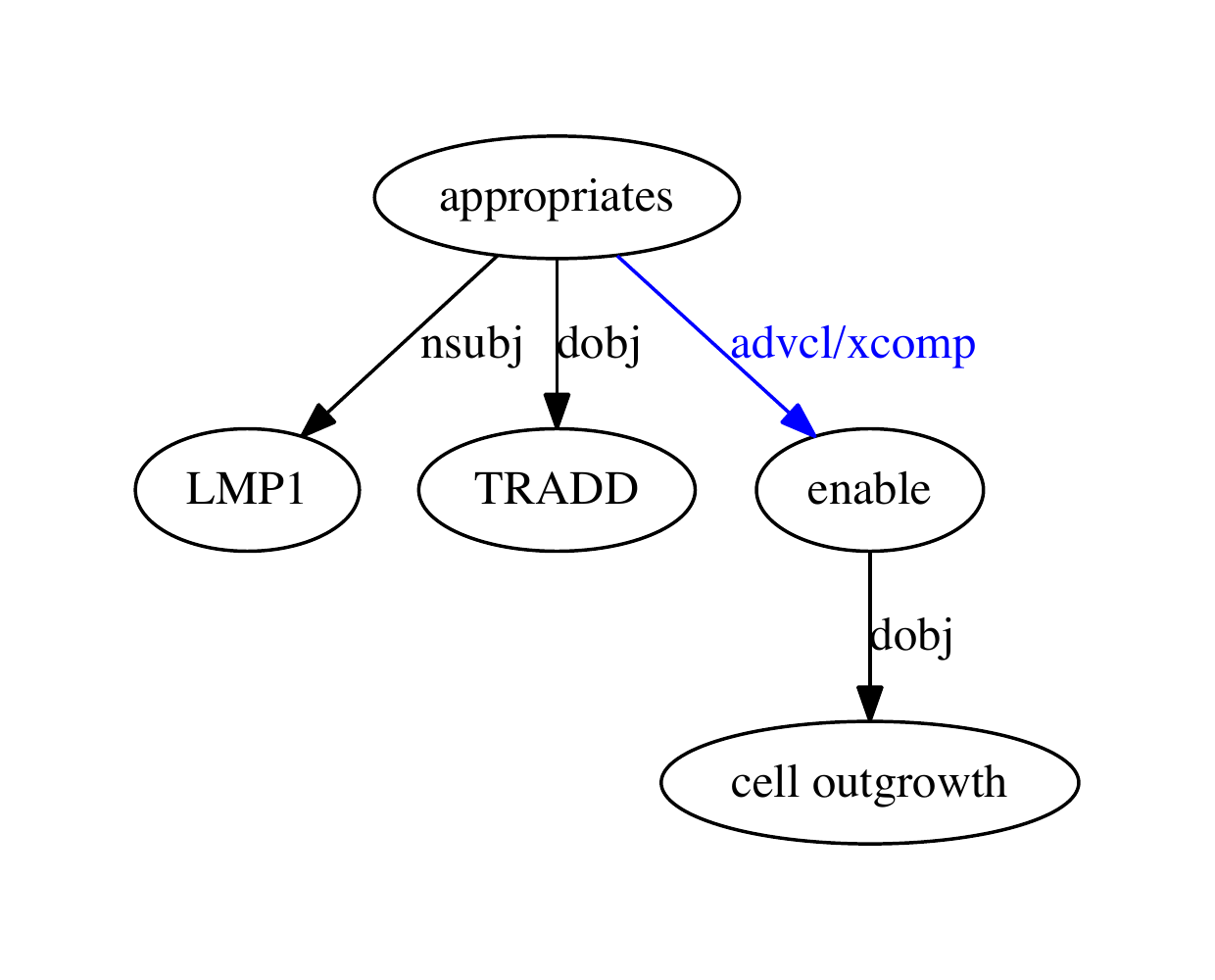}}
\caption{Alternative dependency labels: advcl (Model tree, SNN) or xcomp (BLLIP, Mate).}
\label{fig:advcl_xcomp}
\end{center}
\vskip -0.2in
\end{figure}

In the following three items we look at the reasons of gold standard - test results discrepancies which are not necessarily caused by parser errors. 
\item Dependency variability. 
Sometimes the intended meaning can be expressed in more than one syntactic way. In the following sentence: \textit{These results indicate that LMP1 appropriates TRADD to enable efficient long-term lymphoblastoid cell outgrowth}, gold tree dependency between 'appropriates' and 'enable' is coined 'adverbial clause modifier' ('advcl'), while some parsers identified it as 'external clausal modifier' ('xcomp'). The function of the adverbial clause dependency here is to express purpose. External clause modifier does not have its own subject, but its role is taken over by the object of a higher clause, which is TRADD in our example. In either case we would be able to understand that LMP1 needs TRADD in order to enable cell outgrowth. 'advcl' dependency conveys it directly. The same interpretation is achieved with 'xcomp' but in a compositional way: LMP1 uses TRADD, the latter being necessary for the cell outgrowth (Figure~\ref{fig:advcl_xcomp}). Another example of syntactic variability is 'advcl' $\rightarrow$ 'nmod' (nominal modifier introduced by a preposition) substitutions. In the fragment: \textit{expression of c-myb mRNA is necessary for Hm-induced differentiation}, necessary$\xrightarrow{\text{advcl}}$ differentiation would lead to the same interpretation as necessary$\xrightarrow{\text{nmod:for}}$ differentiation.
\begin{figure}[ht]
\vskip 0.2in
\begin{center}
\centerline{\includegraphics[width=\columnwidth]{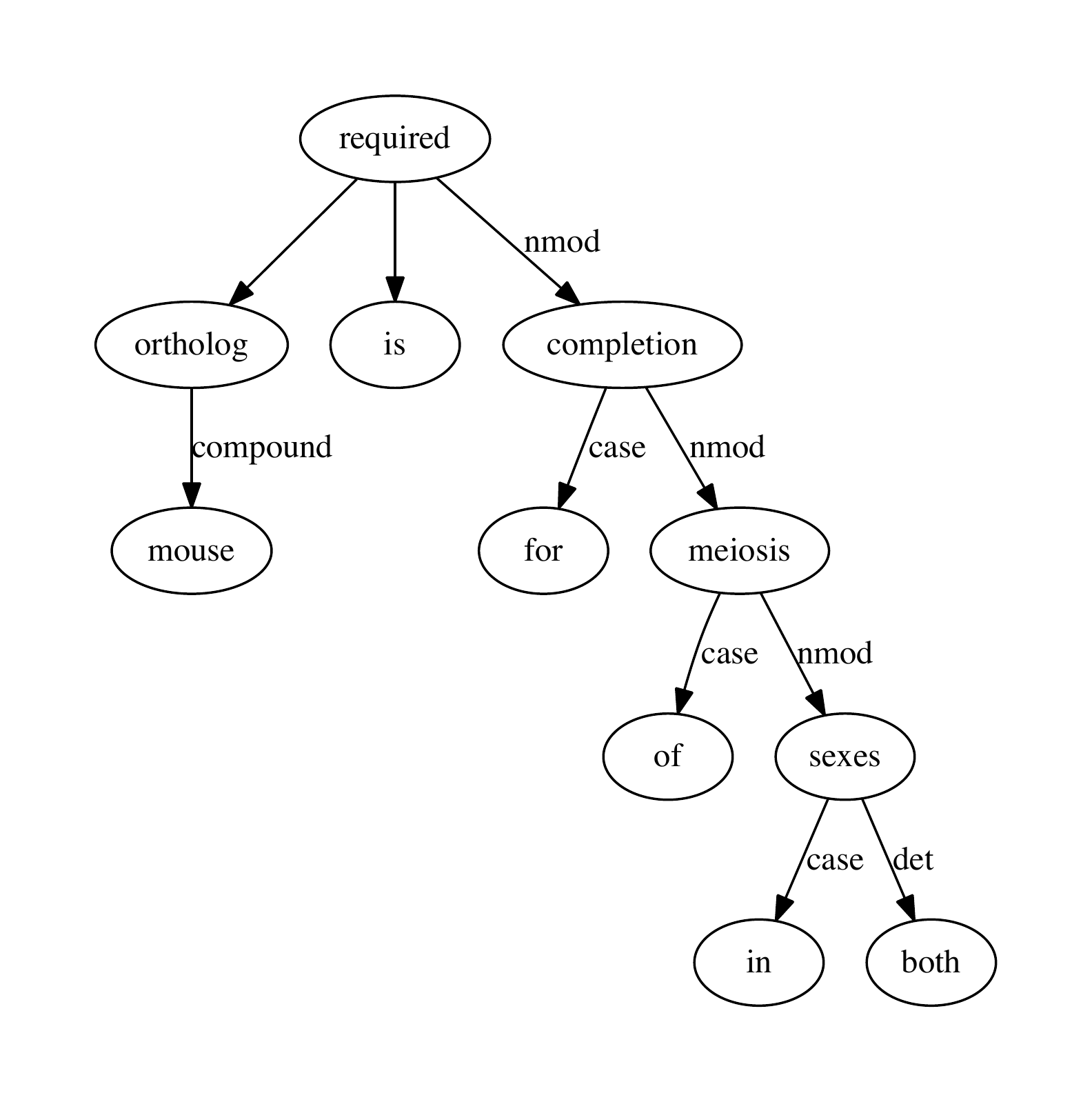}}
\caption{Attachment of a prepositional phrase; serial scheme. (Model tree, Bllip, Mate+Genia model.)}
\label{fig:pp_modified_serial}
\end{center}
\vskip -0.2in
\end{figure}

\begin{figure}[ht]
\vskip 0.2in
\begin{center}
\centerline{\includegraphics[width=\columnwidth]{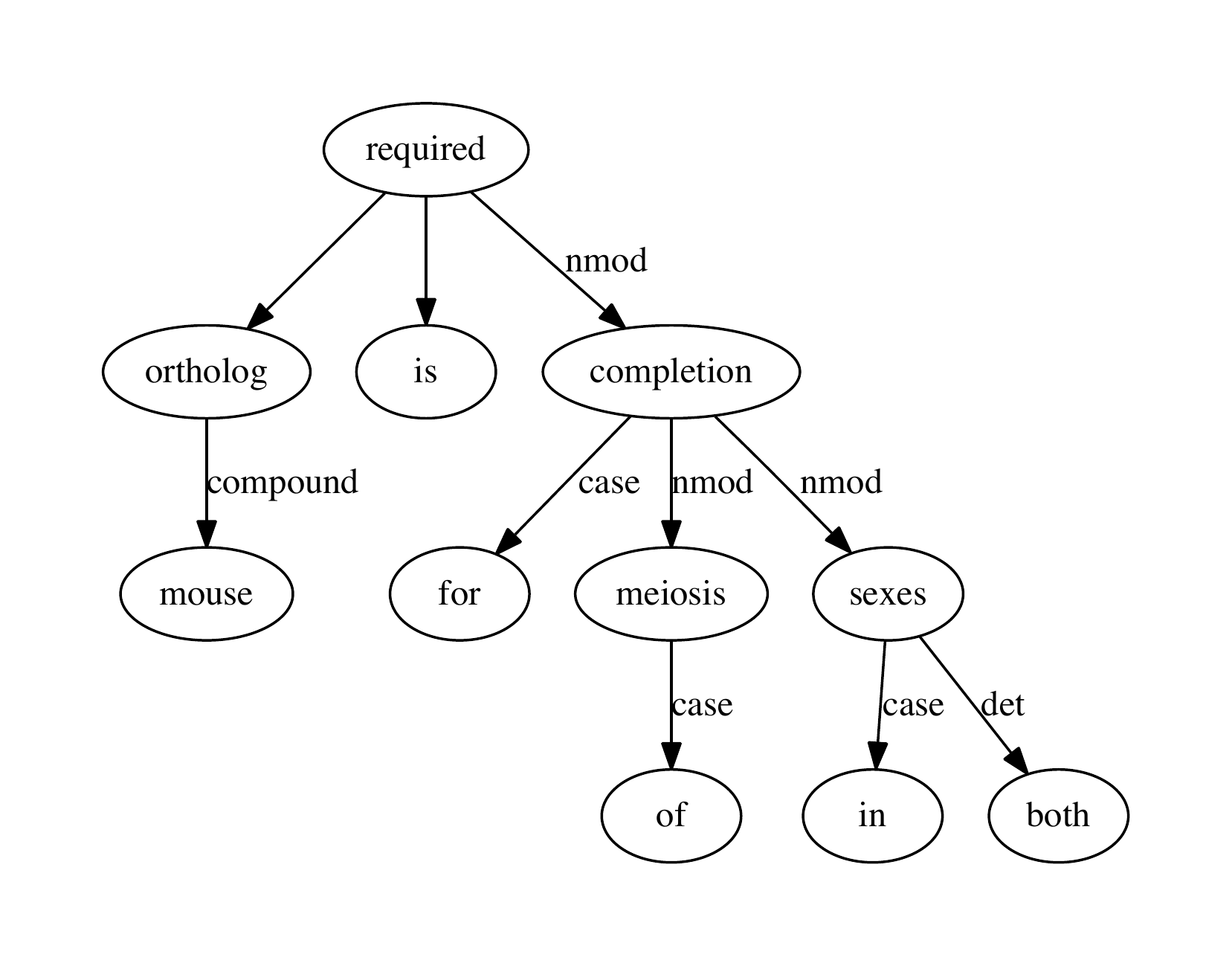}}
\caption{Attachment of a prepositional phrase; parallel scheme. (SNN, Google + Craft and Mixed models.)}
\label{fig:pp_modified_parallel}
\end{center}
\vskip -0.2in
\end{figure}

\item Parsing scheme variability. Similarly to the dependency variability, different ways of attaching constituents lead sometimes to differences between gold and test trees while do not alter the meaning. Consider the following sentence: \textit{We report that the mouse ortholog (Trip13) ... is required for completion of meiosis in both sexes.} The gold tree for the prepositional fragment (Figure~\ref{fig:pp_modified_serial}) suggests serial attachment of prepositional phrases \textit{of meiosis}  and \textit{in both sexes} such that the latter directly modifies the former which in turn is attached to \textit{completion}. Some parsers adopt the same scheme, some others prefer parallel attachment (Figure~\ref{fig:pp_modified_parallel}) in which both prepositional phrases are modifying \textit{completion}. Both trees imply the same interpretation despite topological difference. 
\item Gold standard specificity.
Besides syntactic and parsing schemes flexibility, gold standard annotation itself is not always instructive. Such is the case of 'dep' relation - the most generic one, which only tells that the words are dependent but does not specify, how. A recurrent example from Genia corpus is coordination of premodifiers. Consider the following fragment: \textit{elements with NFkappaB, AP1 and ETS-like binding motifs have been identified...}. The gold annotation suggests 'dep' relation between the motifs $\xrightarrow{\text{dep}}$ NFkappaB. On the contrary, all the parsers labeled the relation as conjunction which seems to be correct. 
\end{itemize}

After having looked closely at parsers treatment of various dependency relations in a diversity of contexts, we conclude that differences between the gold and parse trees may result from parsers errors, input text ambiguity, certain flexibility in choice of the dependency relation, multiple constituent attachment schemes and errors in gold standard annotation.

\section{Final notes: parsing time and recommendations}
\label{sec:time_advise}
\begin{table}[t]
\caption{Parsing time in seconds.}
\label{tab:time}
\begin{tabular}{l|l|c}
\hline
&  Genia-S & Craft-B \\
& (2854)  & (19128) \\
\hline\hline
SNN & 36.4 & 286.9\\\hline
Google &113 & 732.9\\\hline
Mate & 355.8 & 2777\\\hline 
BLLIP &975 & 9832 \\\hline
\end{tabular}
\end{table}

Since parsing is one of the steps in the information extraction pipeline, parser speed matters, especially when large amounts of texts must be processed. Table~\ref{tab:time} shows times required by each of the four top-scored parsers to process GENIA small (2854 sentences) and CRAFT big (19128 sentences) test sets\footnote{All the experiments are performed on 2,9 GHz Intel Core i5 MacBook Pro.}. As one can see, SNN is by far the fastest one, BLLIP - the slowest, Google and Mate are in the middle, with Google being three times faster than Mate. 
Now, with the detailed account of the parsers accuracy and time, we could provide a few recommendations on which parser to choose for information extraction from biomedical articles. If only abstracts are concerned, time is important, and the goal is to accurately identify main semantic roles in terms of subject-predicate-object triplets, SNN is the best option. Google with Genia model provides a respectable alternative for detecting main and context-related dependencies involved in event extraction, especially if time is a concern. If time is not a limiting factor, we would opt for BLLIP as the most accurate parser. 

With regard to full text processing, models trained on CRAFT have obvious advantage. In our experiments, Google performs a little better in identification of main event-related dependencies and patterns involved in the event contextualization. If the choice is to be made between off-the-shelf parsers, BLLIP was more accurate in detection of most of the main event-related dependencies but had serious disadvantage in labeling 'compounds'. Each parser had its stronger and weaker points in labeling contextualization-related dependencies. Therefore, the best strategy here could be a combination of both parsers as long as there are no restrictions on parsing time. 

\section{Conclusions}
In this paper we have studied performance of seven statistical syntactic parsers on biomedical texts. We evaluated the parsers from the point of view of the overall accuracy and identified four best performing ones. The latter have been further analyzed on a selection of syntactic patterns with the task of general and fine-grained event extraction in mind. While interpreting evaluation scores, we pointed out cases, in which formal scores do not pertain to quality of parses.
We have compared portability of various models to abstracts and full texts and found that even within the same genre, models trained on abstracts do not perform respectably on full texts. We have also compared two main gold standards available for biomedical scientific writing, GENIA and CRAFT, and traced possible reasons why parsers trained on abstracts perform poorly on full texts. Finally, we compared the parser speed and provided a few recommendations on parser's choice with respect to the task of event extraction. 

As a future work, we see interest in finding optimal data and technique to train parsers such that they would work equally well on abstracts and full texts. In this light, SNN is attractive as it shows the highest speed. Mate is the one with the lowest time required for training, and thus is also worth exploration. Our experiment with self-training Google parser shows a promising start, and it might be interesting to investigate efficiency of this strategy with different parsers.

\bibliography{ParseEval_biblio}

\begin{thebibliography}{}

\bibitem[Andor et~al.\/, 2016][Andor et~al.\/][2016]{GoogleParser}
Andor, D., Alberti, C., Weiss, D., Severyn, A., Presta, A., Ganchev, K.,
  Petrov, S., \& Collins, M. (2016).
\newblock Globally normalized transition-based neural networks.
\newblock {\em Proceedings of the 54th Annual Meeting of the Association for
  Computational Linguistics, {ACL} 2016, August 7-12, 2016, Berlin, Germany,
  Volume 1: Long Papers}.

\bibitem[Bohnet, 2010][Bohnet][2010]{Mate:Bonnet2010}
Bohnet, B. (2010).
\newblock Top accuracy and fast dependency parsing is not a contradiction.
\newblock {\em {COLING} 2010, 23rd International Conference on Computational
  Linguistics, Proceedings of the Conference, 23-27 August 2010, Beijing,
  China} (pp.\/ 89--97).

\bibitem[Charniak \& Johnson, 2005][Charniak and
  Johnson][2005]{CharniakJohnson2005}
Charniak, E., \& Johnson, M. (2005).
\newblock Coarse-to-fine n-best parsing and maxent discriminative reranking.
\newblock {\em {ACL} 2005, 43rd Annual Meeting of the Association for
  Computational Linguistics, Proceedings of the Conference, 25-30 June 2005,
  University of Michigan, {USA}} (pp.\/ 173--180).

\bibitem[Chen \& Manning, 2014][Chen and Manning][2014]{ChenManningSnn}
Chen, D., \& Manning, C.~D. (2014).
\newblock A fast and accurate dependency parser using neural networks.
\newblock {\em Proceedings of the 2014 Conference on Empirical Methods in
  Natural Language Processing, {EMNLP} 2014, October 25-29, 2014, Doha, Qatar,
  {A} meeting of SIGDAT, a Special Interest Group of the {ACL}} (pp.\/
  740--750).

\bibitem[Chiu et~al.\/, 2016][Chiu et~al.\/][2016]{PyysaloEmbedding}
Chiu, B., Crichton, G., Korhonen, A., \& Pyysalo, S. (2016).
\newblock How to train good word embeddings for biomedical nlp.
\newblock {\em Proceedings of the 15th Workshop on Biomedical Natural Language
  Processing}.

\bibitem[Choi et~al.\/, 2015][Choi et~al.\/][2015]{ItDepends}
Choi, J.~D., Tetreault, J.~R., \& Stent, A. (2015).
\newblock It depends: Dependency parser comparison using {A} web-based
  evaluation tool.
\newblock {\em Proceedings of the 53rd Annual Meeting of the Association for
  Computational Linguistics and the 7th International Joint Conference on
  Natural Language Processing of the Asian Federation of Natural Language
  Processing, {ACL} 2015, July 26-31, 2015, Beijing, China, Volume 1: Long
  Papers} (pp.\/ 387--396).

\bibitem[Clegg \& Shepherd, 2007][Clegg and Shepherd][2007]{Clegg:Shepherd}
Clegg, A.~B., \& Shepherd, A.~J. (2007).
\newblock Benchmarking natural-language parsers for biological applications
  using dependency graphs.
\newblock {\em {BMC} Bioinformatics}, {\em 8}.

\bibitem[Cohen et~al.\/, 2008][Cohen et~al.\/][2008]{Cohen:nominalization}
Cohen, K.~B., Martha, P., \& Lawrence, H. (2008).
\newblock Nominalization and alternations in biomedical language.
\newblock {\em PLoS ONE}, {\em 3}.

\bibitem[Collobert et~al.\/, 2011][Collobert et~al.\/][2011]{Collobert:2011}
Collobert, R., Weston, J., Bottou, L., Karlen, M., Kavukcuoglu, K., \& Kuksa,
  P.~P. (2011).
\newblock Natural language processing (almost) from scratch.
\newblock {\em CoRR}, {\em abs/1103.0398}.

\bibitem[D{\'i}az \& L{\'o}pez, 2015][D{\'i}az and
  L{\'o}pez][2015]{Diaz:Lopez:Tokenization:2015}
D{\'i}az, N. P.~C., \& L{\'o}pez, M. M.~M. (2015).
\newblock An analysis of biomedical tokenization: Problems and strategies.
\newblock {\em Sixth International Workshop on Health Text Mining and
  Information Analysis, 2015, Proceedings} (pp.\/ 40--49).

\bibitem[Kim et~al.\/, 2003][Kim et~al.\/][2003]{GeniaCorpus}
Kim, J., Ohta, T., Tateisi, Y., \& Tsujii, J. (2003).
\newblock {GENIA} corpus - a semantically annotated corpus for bio-textmining.
\newblock {\em Proceedings of the Eleventh International Conference on
  Intelligent Systems for Molecular Biology, June 29 - July 3, 2003, Brisbane,
  Australia} (pp.\/ 180--182).

\bibitem[Klein \& Manning, 2003][Klein and Manning][2003]{KleinManning:2003}
Klein, D., \& Manning, C.~D. (2003).
\newblock Accurate unlexicalized parsing.
\newblock {\em Proceedings of the 41st Annual Meeting of the Association for
  Computational Linguistics, 7-12 July 2003, Sapporo Convention Center,
  Sapporo, Japan.} (pp.\/ 423--430).

\bibitem[Kummerfeld et~al.\/, 2012][Kummerfeld et~al.\/][2012]{ParserShowdown}
Kummerfeld, K.~J., Klein, D., \& Curran, J.~R. (2012).
\newblock Robust conversion of {CCG} derivations to phrase structure trees.
\newblock {\em The 50th Annual Meeting of the Association for Computational
  Linguistics, Proceedings of the Conference, July 8-14, 2012, Jeju Island,
  Korea - Volume 2: Short Papers} (pp.\/ 105--109).

\bibitem[Lease \& Charniak, 2005][Lease and Charniak][2005]{Lease:Charniak}
Lease, M., \& Charniak, E. (2005).
\newblock Parsing biomedical literature.
\newblock {\em Natural Language Processing - {IJCNLP} 2005, Second
  International Joint Conference, Jeju Island, Korea, October 11-13, 2005,
  Proceedings} (pp.\/ 58--69).

\bibitem[McClosky \& Charniak, 2008][McClosky and Charniak][2008]{McClossky}
McClosky, D., \& Charniak, E. (2008).
\newblock Self-training for {Biomedical Parsing}.
\newblock {\em {ACL} 2008, Proceedings of the 46th Annual Meeting of the
  Association for Computational Linguistics, June 15-20, 2008, Columbus, Ohio,
  USA, Short Papers} (pp.\/ 101--104).

\bibitem[McClosky et~al.\/, 2006][McClosky et~al.\/][2006]{SelfTrainPars}
McClosky, D., Charniak, E., \& Johnson, M. (2006).
\newblock Effective self-training for parsing.
\newblock {\em Human Language Technology Conference of the North American
  Chapter of the Association of Computational Linguistics, Proceedings, June
  4-9, 2006, New York, New York, {USA}}.

\bibitem[McClosky et~al.\/, 2010][McClosky et~al.\/][2010]{AnyDomainAdap}
McClosky, D., Charniak, E., \& Johnson, M. (2010).
\newblock Automatic domain adaptation for parsing.
\newblock {\em Human Language Technologies: Conference of the North American
  Chapter of the Association of Computational Linguistics, Proceedings, June
  2-4, 2010, Los Angeles, California, {USA}} (pp.\/ 28--36).

\bibitem[McDonald \& Nivre, 2011][McDonald and Nivre][2011]{McDonaldNivre:2011}
McDonald, R., \& Nivre, J. (2011).
\newblock Analyzing and integrating dependency parsers.
\newblock {\em Computational Linguistics}, {\em 37}, 197--230.

\bibitem[Miyao et~al.\/, 2008][Miyao et~al.\/][2008]{Miyao:2008}
Miyao, Y., S{\ae}tre, R., Sagae, K., Matsuzaki, T., \& Tsujii, J. (2008).
\newblock Task-oriented evaluation of syntactic parsers and their
  representations.
\newblock {\em {ACL} 2008, Proceedings of the 46th Annual Meeting of the
  Association for Computational Linguistics, June 15-20, 2008, Columbus, Ohio,
  {USA}} (pp.\/ 46--54).

\bibitem[Payysalo et~al.\/, 2006][Payysalo et~al.\/][2006]{PayysaloEvaluation}
Payysalo, S., Ginter, F., \& et.al (2006).
\newblock Evaluation of two dependency parsers on biomedical corpus targeted at
  protein-protein interactions.
\newblock {\em International Journal of Medical Informatics}, {\em 75}.

\bibitem[Socher et~al.\/, 2011][Socher et~al.\/][2011]{SocherManning:2011}
Socher, R., Lin, C.~C., Ng, A.~Y., \& Manning, C.~D. (2011).
\newblock Parsing natural scenes and natural language with recursive neural
  networks.
\newblock {\em Proceedings of the 28th International Conference on Machine
  Learning, {ICML} 2011, Bellevue, Washington, USA, June 28 - July 2, 2011}
  (pp.\/ 129--136).

\bibitem[Verspoor et~al.\/, 2012a][Verspoor et~al.\/][2012a]{craft}
Verspoor, K., Cohen, K.~B., \& et.al (2012a).
\newblock A corpus of full-text journal articles is a robust evaluation tool
  for revealing differences in performance of biomedical natural language
  processing tools.
\newblock {\em BMC Bioinformatics}, {\em 13}.

\bibitem[Verspoor et~al.\/, 2012b][Verspoor
  et~al.\/][2012b]{Cohen:Verspoor:2010}
Verspoor, K., Cohen, K.~B., Lanfranchi, A., Warner, C., Johnson, H.~L., Roeder,
  C., Choi, J.~D., Funk, C.~S., Malenkiy, Y., Eckert, M., Xue, N., Jr., W.
  A.~B., Bada, M., Palmer, M., \& Hunter, L.~E. (2012b).
\newblock A corpus of full-text journal articles is a robust evaluation tool
  for revealing differences in performance of biomedical natural language
  processing tools.
\newblock {\em {BMC} Bioinformatics}, {\em 13}, 207.

\end{thebibliography}
\bibliographystyle{mlapa}

\end{document}